\documentclass[letterpaper]{article} 
\usepackage{aaai2026}  
\usepackage{times}  
\usepackage{helvet}  
\usepackage{courier}  
\usepackage[hyphens]{url}  
\usepackage{graphicx} 
\usepackage{natbib}  
\usepackage{caption} 
\usepackage{booktabs}       
\usepackage{amsfonts}       
\usepackage{nicefrac}       
\usepackage{microtype}      
\usepackage{xcolor}         
\usepackage{enumitem} 
\usepackage{amsmath} 
\setlist[itemize]{noitemsep, topsep=0pt}
\usepackage{diagbox} 
\usepackage{float}
\usepackage{tablefootnote}
\usepackage{multirow}
\usepackage{multicol}
\usepackage{adjustbox}
\usepackage{subfig}
\usepackage{graphicx}
\usepackage{caption}
\usepackage{longtable}

\usepackage{graphicx}
\usepackage{makecell}

\usepackage{algorithm}
\usepackage{algorithmicx}
\usepackage{algcompatible} 
\usepackage{algpseudocode}
\usepackage{newfloat}
\usepackage{listings}
\DeclareCaptionStyle{ruled}{labelfont=normalfont,labelsep=colon,strut=off} 
\floatstyle{ruled}
\newfloat{listing}{tb}{lst}{}
\floatname{listing}{Listing}
\title{Optimizing Denoising Trajectories in dLLMs: \\ A Lightweight Evolutionary Heuristic Approach}

\author{
    Zijian Zhao\textsuperscript{\rm 1,2}, Dian Jin\textsuperscript{\rm 3}, Xialiang Tong\textsuperscript{\rm 2}\thanks{Project Lead: Xialiang Tong}, Sen Li\textsuperscript{\rm 1,4}\thanks{Corresponding Author: Sen Li}, Mingxuan Yuan\textsuperscript{\rm 2}\\
}
\affiliations{
    \textsuperscript{\rm 1}The Hong Kong University of Science and Technology\\
    \textsuperscript{\rm 2}Noah's Ark Lab, Huawei\\
    \textsuperscript{\rm 3}The Hong Kong Polytechnic University\\
    \textsuperscript{\rm 4}The Hong Kong University of Science and Technology (Guangzhou)\\

}

\usepackage{bibentry}

\begin{document}

\maketitle

\begin{abstract}
Diffusion Large Language Models (dLLMs) have recently emerged as a promising alternative to conventional Auto-Regressive (AR) Large Language Models (LLMs). By leveraging bidirectional attention and parallel decoding, dLLMs enable more efficient generation. However, they require a carefully designed denoising scheduler at inference time (absent during training) whose choice significantly impacts generation quality. While confidence-based heuristic schedulers have shown strong empirical performance, they suffer from two critical failure modes: EOS Overflow and Proximal Bias. 
Through in-depth analysis of the Transformer's attention patterns, we reveal that these failures stem from certain positions assigning disproportionately high attention weights to invalid tokens (e.g., [MASK] and [EOS]), which produce misleading confidence signals. Building on this insight, empirical evidence shows that valid attention scores can provide complementary guidance to conventional confidence-based heuristics, yet no single metric consistently excels across all scenarios, implying that the optimal denoising trajectory is highly context-dependent.
To address this problem, we propose a lightweight evolutionary heuristic scheduler optimized using the Covariance Matrix Adaptation Evolution Strategy (CMA-ES). Our scheduler dynamically integrates multiple heuristic features with a contextual mean-field embedding, while requiring only 393 trainable parameters. Evaluated on LLaDA and Dream across four reasoning and planning benchmarks, our method consistently outperforms strong baselines, including conventional heuristics, block auto-regressive methods, and recent State-Of-The-Art (SOTA) approaches. To the best of our knowledge, it represents the most parameter-efficient neural scheduler to date.
Our code is available at \url{https://github.com/RS2002/Evo-Denoiser}.
\end{abstract}


\section{Introduction}

Auto-Regressive (AR) Large Language Models (LLMs)~\cite{achiam2023gpt,touvron2023llama,liu2024deepseek} have achieved remarkable success across diverse domains~\cite{chen2026overview}. However, their inherently sequential generation paradigm suffers from slow decoding speed and error accumulation~\cite{arbuzov2025beyond}. In recent years, Diffusion Large Language Models (dLLMs)~\cite{nie2026large,ye2025dream} have emerged as a compelling alternative. By leveraging bidirectional attention and parallel decoding, dLLMs capture full contextual information at every step and enable efficient generation. Recent studies have further demonstrated that discrete dLLMs exhibit scaling laws~\cite{nie2025scaling} and modality expansion capabilities~\cite{you2026llada,zhu2025llada} comparable to those of AR LLMs.

Despite these advantages, dLLMs face a critical train-inference mismatch. During training, dLLMs randomly mask tokens and learn to recover them in parallel under a maximum likelihood Evidence Lower Bound (ELBO) objective~\citep{nie2026large}. At inference, generation starts from a fully masked sequence and proceeds through progressive denoising steps. Since the optimal denoising order is never explicitly supervised, the design of the denoising scheduler becomes crucial. Prior work has shown that the choice of scheduling strategy significantly affects final generation quality~\cite{huang2026reinforcing,he2025mdpo,tang2026your}.

Inspired by confidence-based metrics in AR LLMs (e.g., top-1 probability, entropy, and Gini impurity)~\cite{chen2025first,chen2026confident,kang2026scalable}, recent studies have adopted similar heuristics to guide token denoising in dLLMs~\cite{nie2026large,ben2026accelerated}. Although selecting the top-$k$ most confident tokens substantially outperforms random ordering, confidence-based schedulers still suffer from two persistent failure modes that notably degrade performance, especially in long-form reasoning and planning tasks:
\begin{itemize}[left=0pt]
\item \textbf{EOS Overflow} \cite{kim2026early,kim2026rainbow,park2026confidence}: Excessive EOS tokens accumulate in the rightmost part of the sequence, particularly when they are denoised prematurely.
\item \textbf{Proximal (Local) Bias} \cite{kim2026early,piskorz2026masks}: Once a token is denoised, its neighboring positions receive disproportionately high confidence scores.
\end{itemize}

To address these issues, a growing body of work has focused on improved denoising schedulers, which can be broadly categorized into three types: (i) manually designed heuristics that are computationally efficient and training-free, yet whose optimality is difficult to verify~\cite{cao2026search,park2026confidence}; (ii) Block-AR (Semi-AR) methods that perform parallel denoising within blocks but sequential decoding across blocks~\cite{wu2025fast,zhang2026swordsman}, inherently limiting both inference speed and generation quality; and (iii) trainable schedulers based on SFT or RL that learn an auxiliary network~\cite{hong2026improving,kim2026early,jazbec2025learning,huang2026reinforcing,he2025mdpo}, which are effective but incur substantial training costs (e.g., up to 134M parameters \cite{hong2026improving}). Despite their diversity, these approaches either rely on fixed empirical rules that fail to adapt to varying contexts, or suffer from prohibitive training overhead. This raises a fundamental question: \emph{Is it possible to achieve adaptive, context-aware denoising scheduling with minimal training cost?}

In this paper, we first conduct an in-depth analysis of the Transformer's attention mechanism, revealing that both EOS Overflow and Proximal Bias stem from certain positions excessively attending to invalid tokens such as [MASK] and [EOS]. Building upon this insight, we demonstrate that valid attention scores serve as a strong complementary signal to conventional confidence-based heuristics. Nevertheless, extensive empirical evidence shows that no single heuristic, whether confidence-based or attention-based, consistently dominates, highlighting that the optimal denoising trajectory is highly context-dependent. To address this challenge, we propose a lightweight evolutionary heuristic scheduler optimized using the Covariance Matrix Adaptation Evolution Strategy (CMA-ES)~\cite{hansen2001completely,hansen2016cma}. Our scheduler dynamically integrates multiple heuristic features with a contextual mean-field embedding, while requiring \emph{only 393 trainable parameters}, making it, to the best of our knowledge, the most parameter-efficient neural scheduler to date. Evaluated on LLaDA~\cite{nie2026large} and Dream~\cite{ye2025dream} across four challenging reasoning and planning benchmarks, our method consistently outperforms strong baselines, including conventional heuristics, block-AR methods, and recent State-Of-The-Art (SOTA) approaches, delivering substantial improvements particularly in difficult settings with limited denoising budgets.

\section{Preliminary of dLLMs}
\subsection{Diffusion Language Models}

dLLMs extend diffusion modeling to discrete text generation, offering a compelling alternative to conventional AR LLMs. Unlike AR models that generate tokens sequentially from left to right, dLLMs begin with a fully masked sequence and iteratively denoise tokens in parallel using bidirectional attention. This design enables full-context modeling at every step and provides competitive performance, often with significantly faster inference for long sequences due to parallel decoding. Representative models such as LLaDA~\cite{nie2026large} and Dream~\cite{ye2025dream} have demonstrated strong results while benefiting from inherent parallelism and bidirectional reasoning.

The forward (noising) process gradually corrupts a clean sequence \(\mathbf{x}^0\) by replacing tokens with the special \text{[MASK]} token according to a noise schedule. Let \(\beta_t \in (0,1)\) denote the instantaneous masking rate at time \(t\). The marginal probability that a token remains unmasked at time \(t\) is
\begin{equation}
\alpha_t = \exp\left(-\int_0^t \beta_s \, ds\right),
\end{equation}
which monotonically decreases from $1$ to $0$. The per-token transition kernel is defined as
\begin{equation}
q(x_i^t \mid x_i^{t-1}) =
\begin{cases}
\beta_t & \text{if } x_i^{t-1} \neq \text{[MASK]}, \\
1 & \text{if } x_i^{t-1} = \text{[MASK]},
\end{cases}
\end{equation}
making the masked state absorbing.

The reverse (denoising) process is parameterized by a Transformer network \(p_\theta\), which predicts the original tokens for masked positions. The model is trained by maximizing the ELBO on the data likelihood, which reduces to the following weighted masked cross-entropy objective:
\begin{equation}
\begin{aligned}
&\mathcal{L}_{\text{ELBO}} = \\
& \mathbb{E}_{t \sim \mathcal{U}(0,1),\,\mathbf{x}^0,\,\mathbf{x}^t \sim q}
\left[ \frac{|\dot{\alpha}_t|}{1 - \alpha_t} \sum_{i: x_i^t = \text{[MASK]}} -\log p_\theta(x_i^0 \mid \mathbf{x}^t) \right],
\end{aligned}
\end{equation}
where \(\dot{\alpha}_t = d\alpha_t/dt\).

\subsection{Denoising Scheduler Formulation}

At inference, generation starts from a fully masked sequence \(\mathbf{x}^T\) and progressively recovers tokens over multiple steps. The network \(p_\theta\) estimates the clean data distribution \(p_\theta(\mathbf{x}^0 \mid \mathbf{x}^t)\). Given a noisy sample \(\mathbf{x}^t\), we first draw a predicted clean sequence \(\tilde{\mathbf{x}}^0 \sim p_\theta(\cdot \mid \mathbf{x}^t)\), and then sample the previous state from the posterior:
\begin{equation}
p_\theta(\mathbf{x}_{t-1} \mid \mathbf{x}_t) = \mathbb{E}_{\tilde{\mathbf{x}}^0 \sim p_\theta(\cdot \mid \mathbf{x}^t)} \Bigl[ q(\mathbf{x}_{t-1} \mid \mathbf{x}_t, \tilde{\mathbf{x}}^0) \Bigr].
\label{eq:denoise}
\end{equation}
However, computing $p_\theta(\mathbf{x}_{t-1} \mid \mathbf{x}_t)$ directly is computationally intractable due to the expectation over the vocabulary space. In practice, we mostly rely on a denoising scheduler to approximate this reverse step. Importantly, the inference-time denoising schedule (e.g., the number of iterations and the re-masking strategy) is not explicitly supervised during training, as the ELBO objective jointly supervises all time steps. Recent studies \cite{tang2026your} have shown that the choice of this schedule significantly affects generation quality, making the design of effective denoising schedulers a critical challenge for dLLMs. A detailed related works review is provided at Appendix.

\section{Failure Analysis of Heuristic Schedulers}

In this section, we analyze the top-1 probability denoising scheduler as a representative example to illustrate the underlying causes of EOS Overflow and Proximal Bias through the lens of the Transformer's attention mechanism. We also provide both empirical and intuitive evidence demonstrating that conventional single-token heuristic metrics are insufficient for determining the optimal denoising strategy.

\subsection{Why Confidence Is Not Enough}

We first formally define the heuristic scheduler for dLLM denoising. At inference time, generation begins from a fully masked sequence \(x_T = [\text{[MASK]}]^L\), where \(L\) is the sequence length (the prompt preceding \(x_T\) is omitted for simplicity). At each denoising step \(t\), let \(\mathcal{M}_t\) denote the set of positions that remain masked:
\begin{equation}
\mathcal{M}_t = \{i \mid x_{t,i} = \text{[MASK]}\}.
\end{equation}

The dLLM produces intermediate representations \(\text{f}_{\theta}(x_t)\) (e.g., hidden states or logits), which are passed through a scorer \(\text{h}(\cdot)\) to obtain denoising scores:
\begin{equation}
s_{t,i} = \text{h}(\text{f}_{\theta}(x_{t,i})),
\label{eq:scorer}
\end{equation}
where \(\text{h}(\cdot)\) extracts features such as top-1 probability, entropy, or margin probability. The top-\(k\) highest-scoring masked tokens are then selected for denoising:
\begin{equation}
\mathcal{I}_t = \operatorname{Top\text{-}K}\bigl( s_{t,i} \mid i \in \mathcal{M}_t \bigr).
\label{eq:topk}
\end{equation}

We use the top-1 probability heuristic as a case study to uncover the root causes of EOS Overflow and Proximal Bias, which have not been thoroughly explained in prior work. Since dLLMs are built upon the Transformer architecture, we analyze their behavior through attention patterns. At step \(t\), for the \(m\)-th attention head in the \(n\)-th layer, the attention matrix is:
\begin{equation}
A^{n,m}_t = \mathrm{Softmax}\left( \frac{Q^{n,m}_t (K^{n,m}_t)^\top}{\sqrt{d_k}} \right),
\end{equation}
where \(Q^{n,m}_t\) and \(K^{n,m}_t\) are the query and key matrices, respectively, and \(d_k\) is the head dimension.

Unlike AR LLMs, dLLM sequences contain many [MASK] tokens and trailing [EOS] tokens, both of which often receive uninformative attention. We therefore define the {valid attention score} for each token as:
\begin{equation}
a^n_{t,i} = \frac{1}{H} \sum_{m=1}^H \sum_{j} A^{n,m}_t[i,j] \cdot \mathbf{1}\{x_{t,j} \in \mathcal{V}\},
\end{equation}
where \(\mathcal{V}\) is the set of valid (non-[MASK], non-[EOS]) vocabulary tokens and \(H\) is the number of attention heads.

\begin{figure}[t!]
\centering
\subfloat[Step 0]{\includegraphics[width=0.5\textwidth]{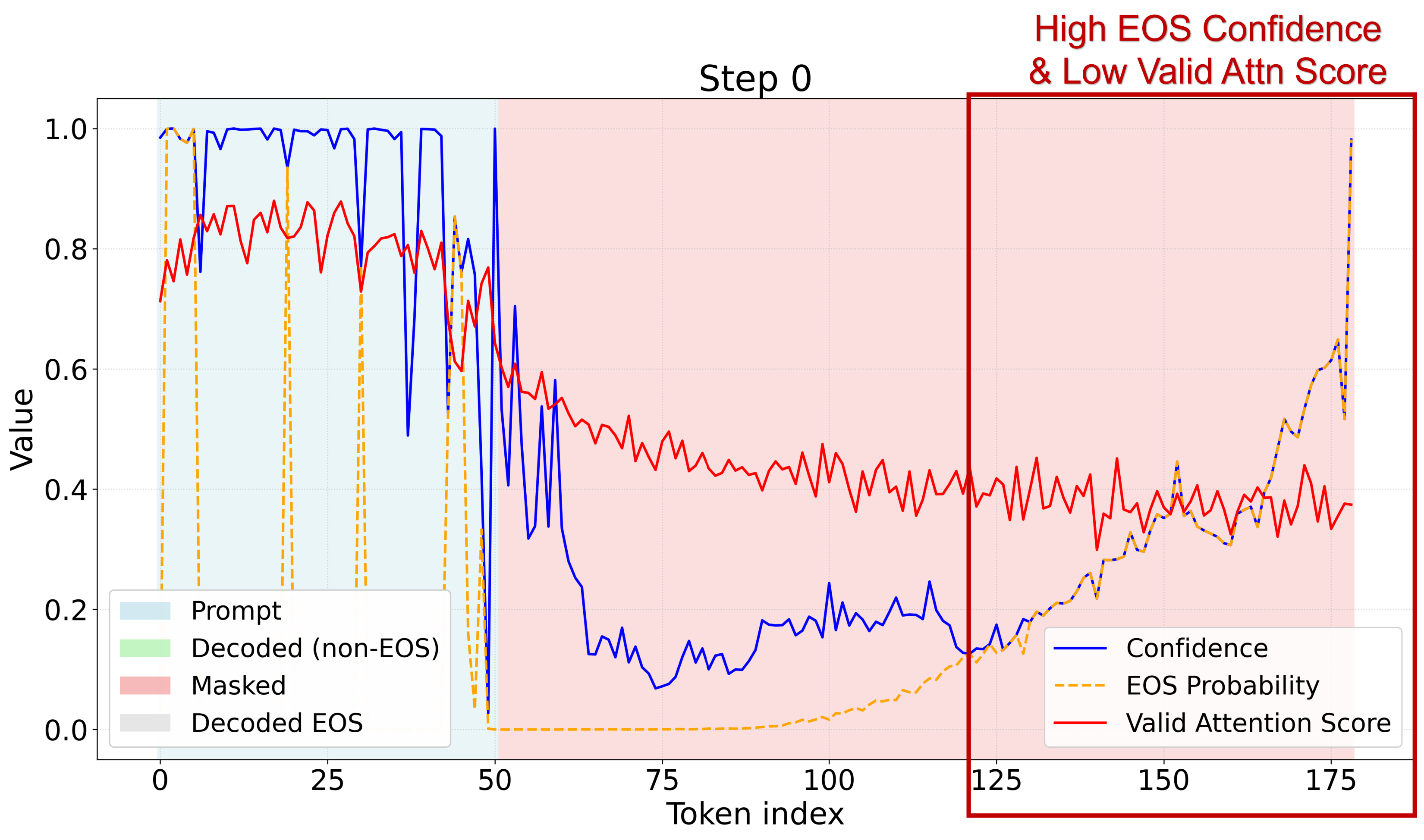}} \\
\subfloat[Step 3]{\includegraphics[width=0.5\textwidth]{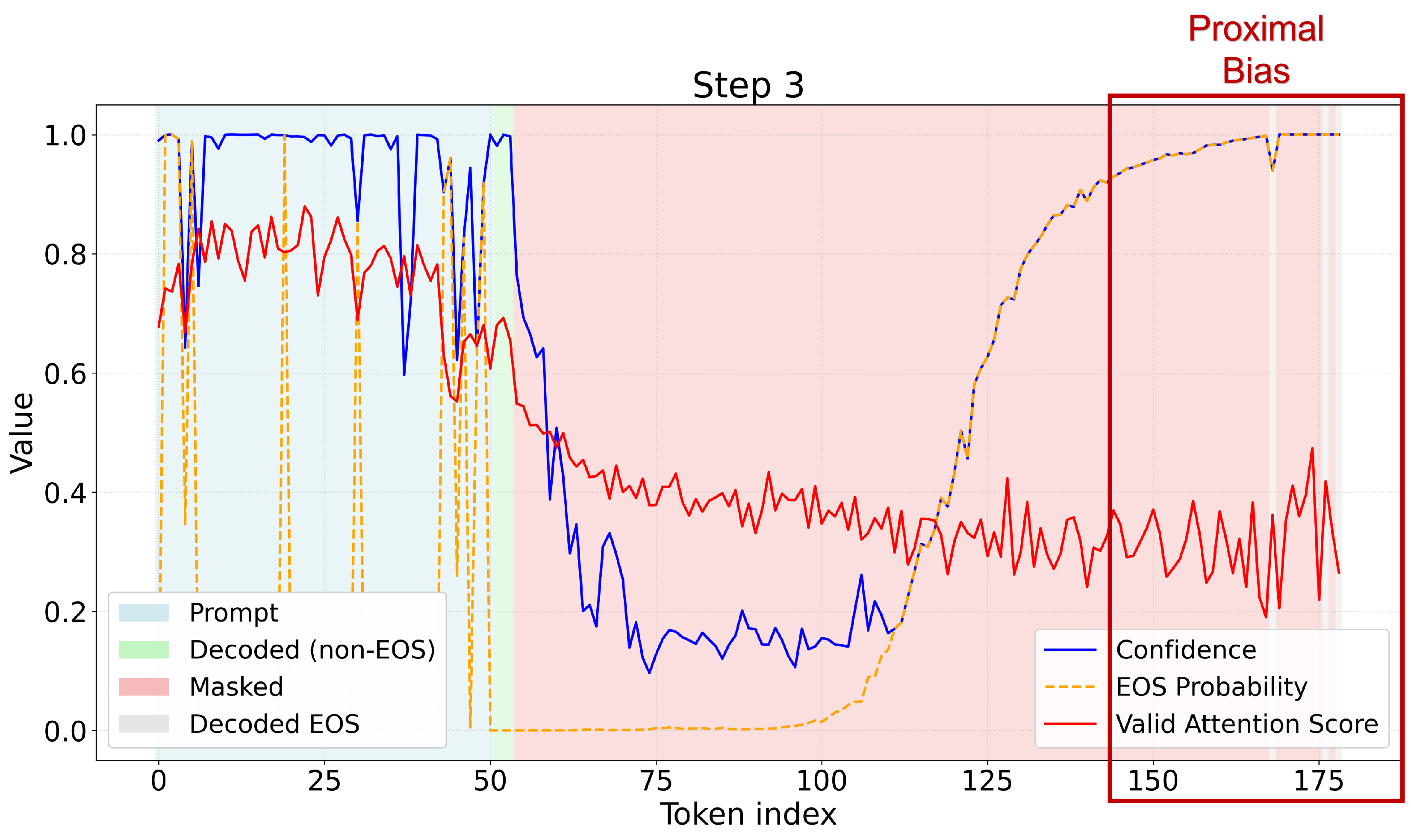}} \\
\subfloat[Step 56]{\includegraphics[width=0.5\textwidth]{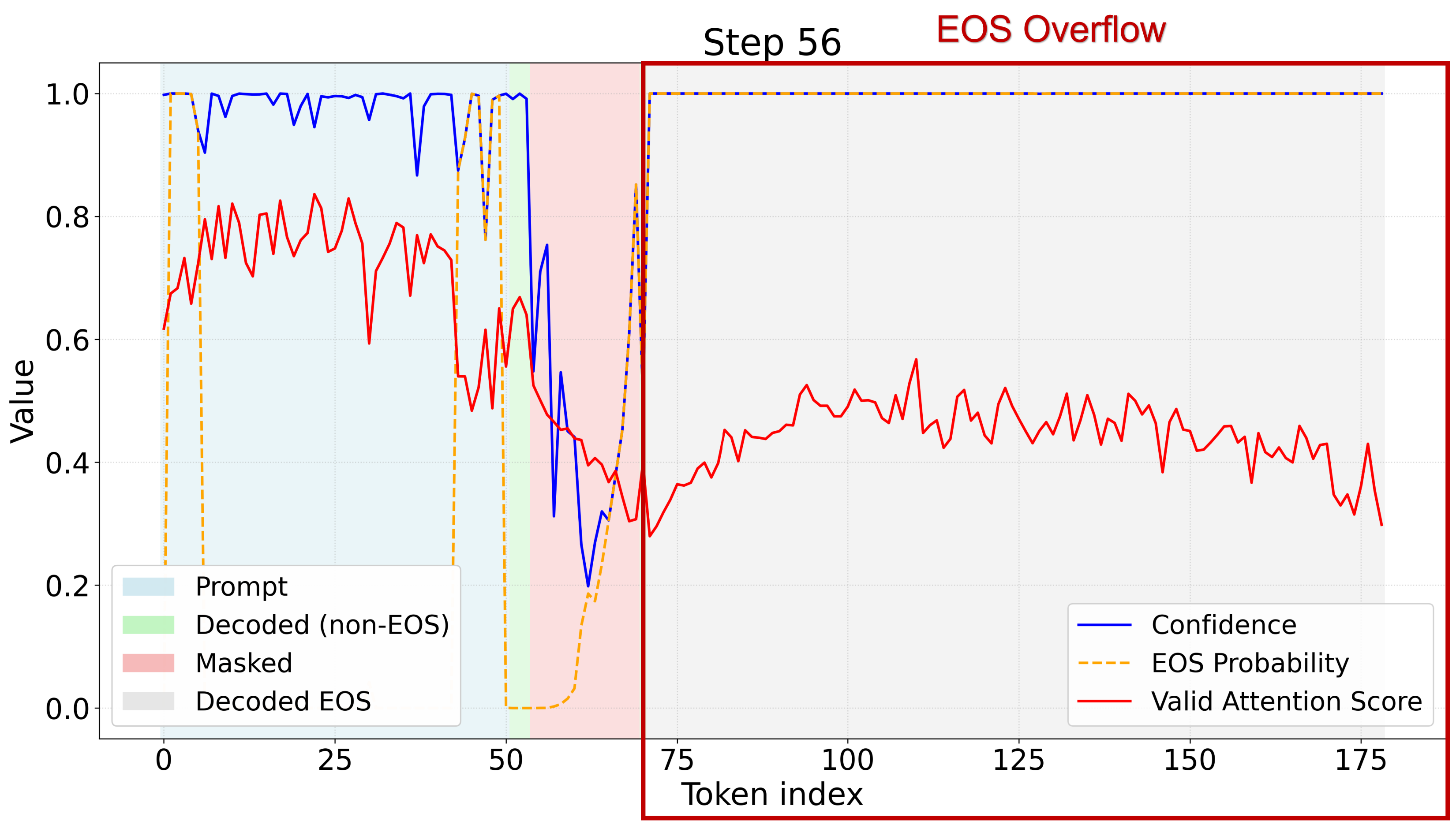}} \\
\caption{Denoising process visualization on a GSM8K example using LLaDA-8B-Instruct. Different background colors represent prompt (blue), [MASK] (red), and [EOS] (gray) regions. The curves show top-1 probability (blue), valid attention score (red), and EOS probability (yellow).}
\label{fig:process}
\end{figure}

Fig.~\ref{fig:process} visualizes the denoising process on a GSM8K sample (here, step \(i\) corresponds to time step \(T-i\) for readability). The valid attention scores are computed from the middle layer as a representative example. We observe the following:
\begin{itemize}[left=0pt]
\item \textbf{Step 0 (initial):} Both beginning and ending tokens exhibit high top-1 probabilities. While high confidence at the beginning is expected due to strong prompt context, high confidence at the end is surprising. These ending tokens show low valid attention scores but high EOS prediction probabilities, indicating that they assign excessive attention to uninformative [MASK] tokens, resulting in misleadingly high confidence.
\item \textbf{Step 3:} Several ending tokens have been decoded as [EOS]. Neighboring tokens then exhibit a sharp increase in both top-1 probability and EOS probability, which is a clear manifestation of \textbf{Proximal Bias}. Their valid attention scores decrease compared to Step 0, suggesting increased attention to the newly decoded (but uninformative) [EOS] tokens.
\item \textbf{Step 56:} From Step 3 to Step 56, only ending tokens continue to be decoded as [EOS], while beginning tokens remain largely unchanged. This illustrates the \textbf{EOS Overflow} phenomenon, where [EOS] tokens occupy an excessive portion of the sequence.
\end{itemize}

These observations reveal that both failure modes originate from the valid attention mechanism: ending tokens initially over-attend to [MASK] tokens and decode prematurely as [EOS]; the newly generated [EOS] tokens then attract further attention, causing neighboring tokens to follow suit. This cascading effect leads to severe Proximal Bias and EOS Overflow. Crucially, these findings suggest that a purely confidence-based scheduler is fundamentally limited: it cannot distinguish between genuine predictive certainty and attention-driven false confidence, motivating our exploration of alternative signals.

\subsection{Why We Need Context with Multiple Heuristics}

\begin{figure}[t!]
\centering
\includegraphics[width=0.48\textwidth]{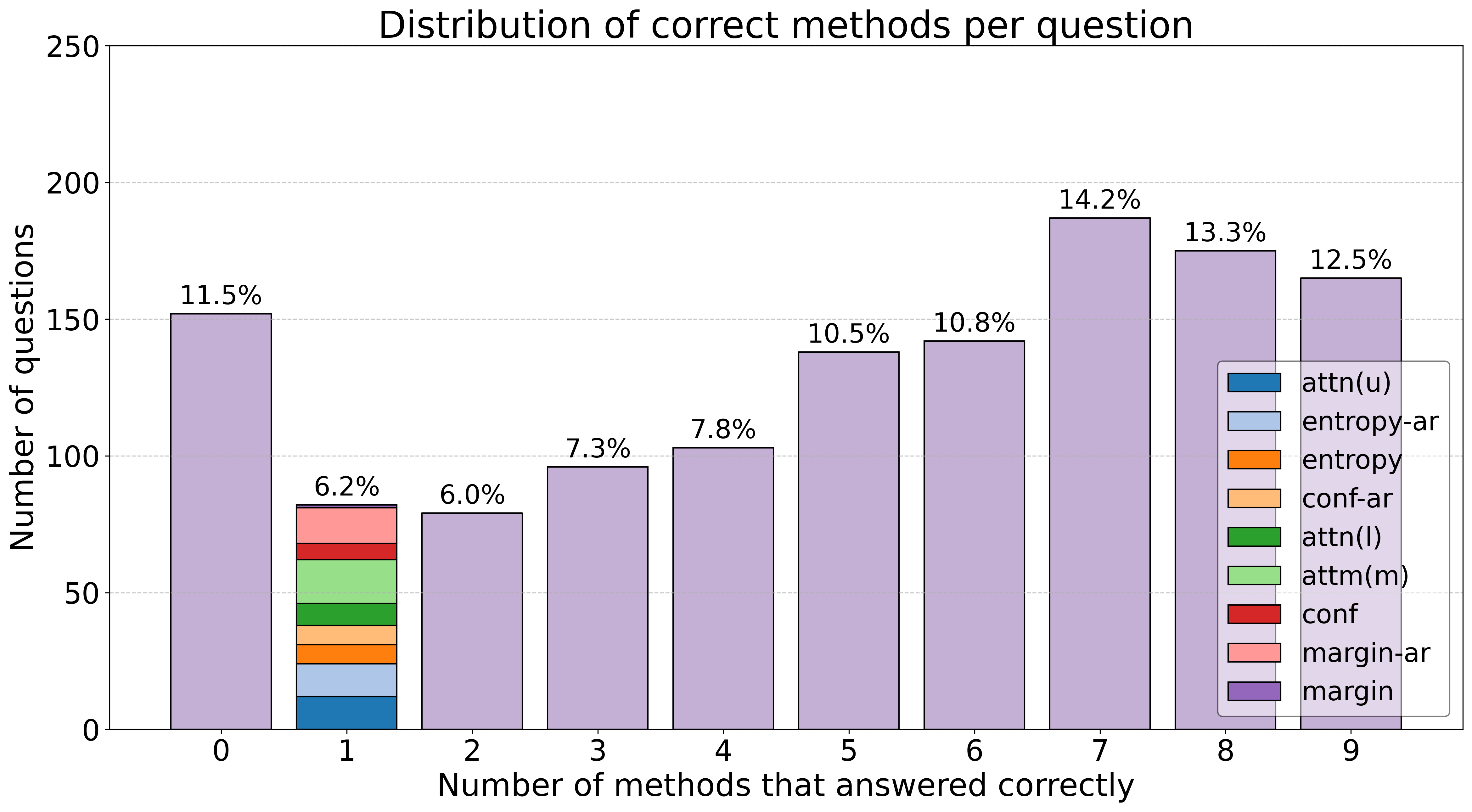}
\caption{Distribution of correctly answered questions by different heuristics on GSM8K (generation length 128, denoising budget 32). Each bar shows the number of questions solved by individual or combined heuristics.}
\label{fig:analysis}
\end{figure}

To further investigate the limitations of single-heuristic schedulers, we evaluate several confidence-based methods on GSM8K with a generation length of 128 and a denoising budget of 32. The tested heuristics include top-1 probability, entropy, margin probability, and their Block-AR variants (block size 32). Motivated by our attention analysis, we also examine valid attention scores computed separately from upper, middle, and lower layers (inspired by prior findings that different layers capture distinct syntactic and semantic information~\cite{vig-2019-multiscale,jawahar2019does}).

Fig.~\ref{fig:analysis} shows the number of correctly answered questions for each scheduler, with special emphasis on questions solved uniquely by one method. Detailed results are provided in Table~\ref{tab:llada}. Our analysis yields several key insights:

\begin{itemize}[left=0pt]
\item Although individual scheduler accuracies range from 46.2\% to 66.6\% (Table~\ref{tab:llada}), only 11.5\% of questions are answered incorrectly by \emph{all} schedulers (Fig.~\ref{fig:analysis}). This indicates that dLLMs possess significantly stronger problem-solving capability than current schedulers can elicit. In other words, if an oracle scheduler existed, the upper-bound performance would be at least 88.5\%, which is comparable to or even exceeds some post-training results via SFT or RL\cite{zhao2026d1,xie2026advancing}. This highlights the critical importance of optimizing the denoising trajectory.
\item Each heuristic exhibits unique strengths: certain questions are solved exclusively by one scheduler but not others. This leads to two important conclusions: (i) no single manually designed heuristic is universally sufficient, and combining multiple complementary heuristics is required; (ii) the optimal denoising strategy is highly context-dependent, as no single fixed heuristic is suitable for all questions.
\end{itemize}

To further illustrate why context matters, consider the interplay between attention and confidence: a high valid attention score may indicate sufficient information flow, but alone it cannot determine whether the token is ready to be denoised. However, when accompanied by a high top-1 probability, this combined signal provides stronger evidence. Moreover, the overall magnitude of attention scores varies with sequence length, as longer prompts naturally provide richer valid contexts. This observation underscores the necessity of context-aware scheduling: the decision to denoise a token cannot be made in isolation but must account for the global state of the sequence.

\section{Methodology}

\begin{figure}[t!]
\centering
\includegraphics[width=0.48\textwidth]{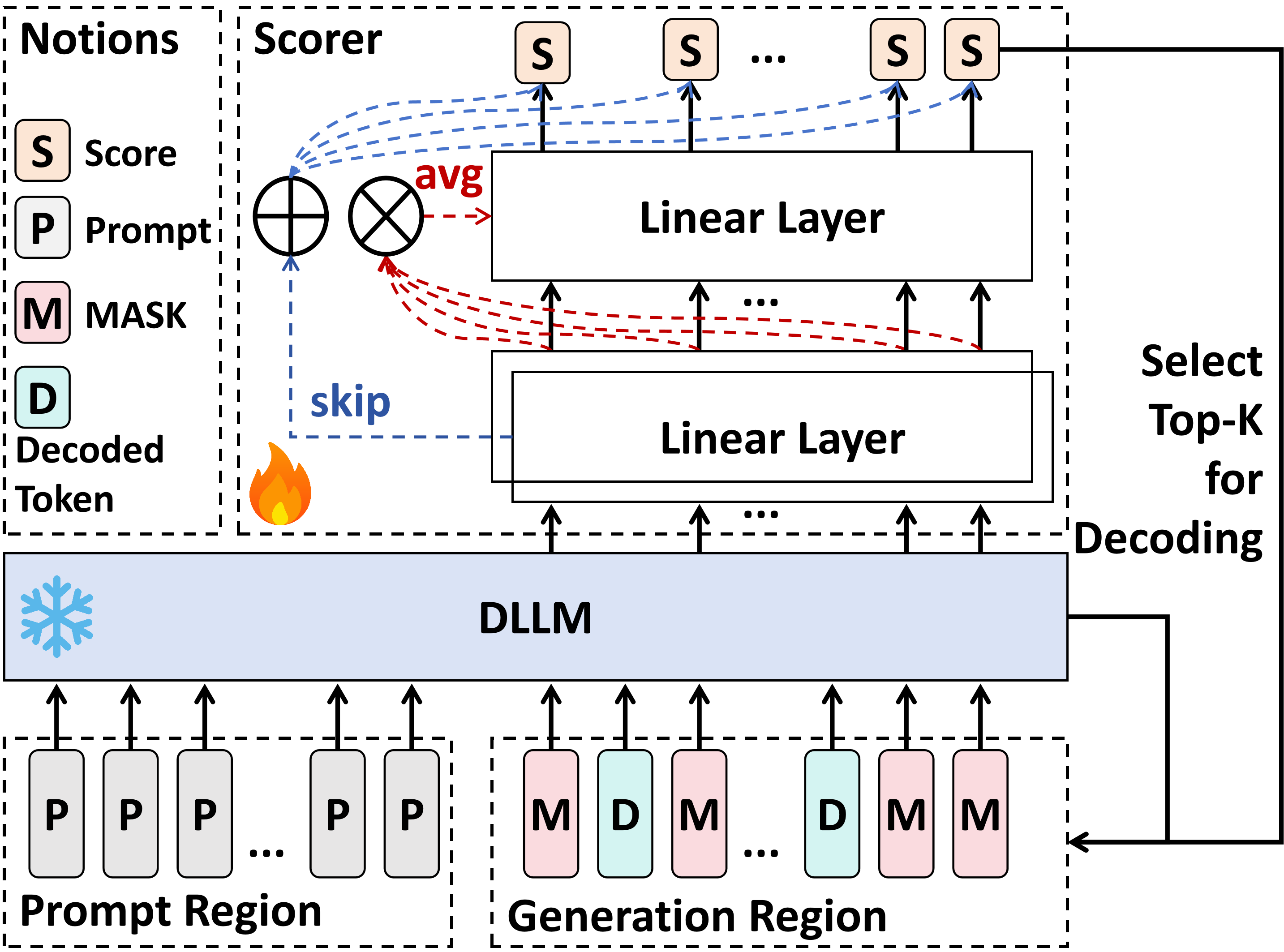}
\caption{Network architecture of the proposed neural scorer.}
\label{fig:main}
\end{figure}

Motivated by the insights from our failure analysis, we propose an evolutionary heuristic scheduler that learns to dynamically combine multiple heuristics for superior denoising quality. Following the formulation in Eq.~\eqref{eq:scorer}, we parameterize the scorer as \(\mathrm{h}_{\Theta}(\cdot)\). Directly optimizing the scorer parameters via policy-gradient methods (e.g., PPO~\cite{schulman2017proximal}, GRPO~\cite{shao2024deepseekmath}) is prohibitively difficult: the token selection operation in Eq.~\eqref{eq:topk} is discrete and non-differentiable, and backpropagating through the entire dLLM is computationally infeasible. We therefore reformulate scheduler optimization as a black-box problem and adopt CMA-ES~\cite{hansen2001completely,hansen2016cma}, a zero-order evolutionary strategy that only requires evaluating the final task accuracy as the fitness signal. Although our method does not explicitly model long-term future effects through the Bellman equation as in RL-based schedulers~\cite{huang2026dont,hong2026improving}, it is not myopic: because the scorer is shared across all denoising steps, optimizing for final answer accuracy implicitly captures temporal dependencies among successive decisions.

\subsection{Feature Construction}

We formalize the input features for our evolutionary scorer. The manual heuristic features are as follows:

\begin{itemize}[left=0pt]
\item \textbf{Top-1 Probability}: The predicted probability of the most likely token:
\begin{equation}
\text{Top-1 Prob} = \max p_{\theta}(\cdot \mid x_{t,i}).
\end{equation}

\item \textbf{Probabilistic Margin}: The difference between the top-1 and top-2 probabilities, which effectively quantifies prediction confidence:
\begin{equation}
\text{Prob Margin} = \max p_{\theta}(\cdot \mid x_{t,i}) - \max^{(2)} p_{\theta}(\cdot \mid x_{t,i}),
\end{equation}
where \(\max^{(2)}\) denotes the second-highest probability.

\item \textbf{Valid Attention Score}: We extract attention scores from three representative layers, including layer \(0\) (bottom), \(\lfloor l/2 \rfloor\) (middle), and \(l-1\) (top), where \(l\) is the total number of layers in the dLLM backbone. This choice is motivated by prior findings that different layers capture distinct levels of syntactic and semantic information \cite{vig-2019-multiscale,jawahar2019does}.
\end{itemize}

Additionally, inspired by the success of Block-AR methods, we incorporate positional information. We include the relative position \(\frac{i}{L}\) and the relative position among currently masked tokens:
\begin{equation}
\text{MASK-Pos} = \frac{\sum_{j \in \mathcal{M}_t} \mathbf{1}\{j < i\}}{|\mathcal{M}_t|}.
\end{equation}

These features together form a 7-dimensional vector \(\kappa_{t,i} \in [0,1]^7\) for each token at step \(t\) (1 for top-1 probability, 1 for margin, 3 for attention from different layers, 1 for relative position, and 1 for mask position). In our design, all features have the same range, which is beneficial for training. While this design results from careful empirical selection, future work may explore additional heuristics such as entropy or alternative attention-based signals~\cite{guo2024attention}.

\subsection{Network Architecture}

The architecture of our neural scorer is illustrated in Fig.~\ref{fig:main}. It employs a dual-path design: a linear skip-connection path (blue) and a two-layer Mean-Field Multilayer Perceptron (MF-MLP) path (red). The final score is computed as:
\begin{equation}
s_{t,i} = \mathrm{Linear}(\kappa_{t,i}) + \mathrm{MF\text{-}MLP}(\kappa_{t,i}; \kappa_t).
\end{equation}

This design is motivated by two key observations from our analysis. First, since the optimal denoising strategy is context-dependent, the scorer must incorporate global contextual information. However, full Transformer-based scorers \cite{hong2026improving,jazbec2025learning} introduce too many parameters for effective zero-order optimization. Inspired by Mean-Field Reinforcement Learning (MFRL) \cite{yang2018mean}, we propose a lightweight MF-MLP that uses the mean embedding of the first-layer hidden representations as contextual input to the second layer. We apply the mean-field operation only after the first layer to minimize parameter count. Second, a purely non-linear architecture can slow down early-stage optimization. Since individual heuristics already perform reasonably well in many cases, we add a linear bypass path to enable fast initial progress via simple weighted combinations.

Overall, this design reflects our goal of achieving a compact yet effective structure. To capture the mean-field contextual information, a single layer is insufficient, as an affine combination would be identical for all tokens. Thus, we adopt a two-layer MF-MLP, with the mean-field embedding introduced only at the second-layer input to minimize the parameter overhead. Additionally, the linear bypass layer facilitates efficient training under zero-order optimization conditions.

We now formalize the MF-MLP structure. First, a linear layer with ReLU activation and layer normalization produces per-token embeddings:
\begin{equation}
y_{t,i} = \mathrm{Linear}(\kappa_{t,i}) \in \mathbb{R}^h,
\end{equation}
where \(h\) is the hidden dimension. The mean-field embedding over all masked tokens is then computed:
\begin{equation}
\overline{y}_t = \frac{1}{|\mathcal{M}_t|} \sum_{j \in \mathcal{M}_t} y_{t,j} \in \mathbb{R}^h.
\end{equation}
Finally, the second linear layer produces the contextual score:
\begin{equation}
z_{t,i} = \mathrm{Linear}([y_{t,i}; \overline{y}_t]) \in \mathbb{R}.
\end{equation}
The final score \(s_{t,i}\) is the sum of \(z_{t,i}\) and the affine bypass term.

\subsection{Training Process}

We employ CMA-ES with task-specific adaptations for training. The fitness function is defined as the average accuracy over a mini-batch:
\begin{equation}
f(\Theta) = \frac{1}{B} \sum_{q \in \mathcal{B}} \mathbf{1}\{\text{correct}(q; \mathrm{h}_{\Theta})\},
\end{equation}
where \(\mathcal{B}\) is a mini-batch of size \(B\) sampled from the training set. The batch is fixed within each generation but resampled across generations to reduce overfitting. The complete training procedure is summarized in Algorithm~\ref{alg:cmaes} at Appendix. Since zero-order optimization itself is not the primary contribution of this work, we refer readers to \cite{hansen2001completely,hansen2016cma} for detailed CMA-ES mechanics.

\begin{table*}[t!]
\centering
\caption{Model Performance on LLaDA-8B-Instruct \cite{nie2026large}. In the table, `T' denotes the step budget, `L' denotes the sequence length, `B' denotes the block size, and `Attn (u/m/l)' indicates that the attention score is computed from the upper, middle, or lower layer, respectively. \textbf{Bold} indicates the best performance and \underline{underline} denotes the second best one. These conventions apply to all subsequent tables. The results reported for EDM are reproduced from the original paper \cite{kim2026early}.}
\centering
\begin{tabular}{lllcccccccc}
\toprule
& & \multicolumn{4}{c}{GSM8K} & \multicolumn{4}{c}{Math} \\
\cmidrule(lr){3-6} \cmidrule(lr){7-10}
\multirow{2}{*}{Type} & \multirow{2}{*}{Scheduler} & \multicolumn{2}{c}{L=128} & \multicolumn{2}{c}{L=256} & \multicolumn{2}{c}{L=128} & \multicolumn{2}{c}{L=256} \\
\cmidrule(lr){3-4} \cmidrule(lr){5-6} \cmidrule(lr){7-8} \cmidrule(lr){9-10}
& & T=16 & T=32 & T=16 & T=32 & T=16 & T=32 & T=16 & T=32 \\
\midrule
\multirow{3}{*}{Heuristics} & Top-1 Prob & 43.2 & 54.0 &40.4 &47.5 & 17.8& 21.0 & 13.2& 18.6\\
& Entropy & 35.9 & 46.2 & 43.2 & 44.0 & 17.2 & 13.0 &12.2 & 15.2\\
& Prob Margin & 50.2 & 54.9 & 45.6 & 50.0 & \underline{19.8} & 20.0 & \underline{17.2} & \underline{19.8} \\
\midrule
\multirow{3}{*}{Block-AR (B=32)} & Top-1 Prob & 42.6 & 60.5 & 9.8 & 46.8 & 11.8 & 23.0 & 4.8 & 17.4 \\
& Entropy & 32.8 & 57.0 & 6.7 & 35.1 & 11.8 & 19.2 & 4.0 & 11.2 \\
& Prob Margin & 47.6 & \underline{66.6} & 12.8 & 52.5 & 15.2 & \underline{24.0} & 5.4 & 16.8 \\
\midrule
\multirow{4}{*}{Recent Works} & EDM (5M params) & -- & 56.8 &  -- &  -- & -- & 22.8 &  -- &  -- \\
& CCD & 43.5 & 52.5 & 37.0 & 48.7  & 18.2 & 19.4 &  12.2 & 17.4 \\
& AGDO Scheduler  & 43.4 & 54.4 & 40.3 & 47.4 & 17.6 & 21.4 & 13.2& 19.0 \\
& Suffix Anchor & \underline{54.9} & 56.7 & \underline{45.7} & \underline{50.9} & 18.0 & 21.8  & 16.8 & 19.2 \\
\midrule
\multirow{4}{*}{Proposed} & Attn (u) & 32.5 & 53.5 & 9.6 & 33.2 & 5.8 & 18.0 &6.4 &11.8 \\
& Attn (m) & 38.2 & 59.1 & 40.2 & 50.8 & 10.2 & 17.8 & 12.4 & 17.2 \\
& Attn (l) & 23.6 & 54.2 & 5.6 & 30.5 & 4.6 & 17.0 & 2.6 & 7.2 \\
& Evolution (393 params) & \textbf{58.5} & \textbf{67.6} & \textbf{52.6} & \textbf{60.8} & \textbf{21.4} & \textbf{27.2} & \textbf{25.0} & \textbf{27.8} \\
\bottomrule
\toprule
& & \multicolumn{4}{c}{Countdown} & \multicolumn{4}{c}{StrategyQA} \\
\cmidrule(lr){3-6} \cmidrule(lr){7-10}
\multirow{2}{*}{Type} & \multirow{2}{*}{Scheduler} & \multicolumn{2}{c}{L=128} & \multicolumn{2}{c}{L=256} & \multicolumn{2}{c}{L=128} & \multicolumn{2}{c}{L=256} \\
\cmidrule(lr){3-4} \cmidrule(lr){5-6} \cmidrule(lr){7-8} \cmidrule(lr){9-10}
& & T=16 & T=32 & T=16 & T=32 & T=16 & T=32 & T=16 & T=32 \\
\midrule
\multirow{3}{*}{Heuristics} & Top-1 Prob & 39.8 & 46.0 & 3.5 & 20.7 & 63.0 & 65.9 & 36.8 & 56.8 \\
& Entropy & 39.5 & 40.6  & 7.4 & 27.0 & 62.7 & 65.5 & 33.5 & 51.4 \\
& Prob Margin & 40.3 & \underline{48.4} & 6.6 & 18.8 & 63.2 & 65.9 & 41.6  & 57.9 \\
\midrule
\multirow{3}{*}{Block-AR (B=32)} & Top-1 Prob & 24.2 & 36.3  & 4.7 & 9.0 &55.0 & 62.9 & 26.9 & 42.4 \\
& Entropy &  16.8 & 28.1 &  3.5 & 9.8 & 47.6  & 61.6 & 27.4 & 44.0 \\
& Prob Margin & 26.6 & 39.8 & 3.5 & 11.3 & 53.1 & 62.6 & 23.0 & 40.6 \\
\midrule
\multirow{4}{*}{Recent Works} & EDM (5M params) & -- & 43.8 &  -- &  -- & -- & -- &  -- &   -- \\
& CCD & 37.9 & 43.0 & 4.7 & 17.6& \underline{63.9} & \textbf{66.4} & 30.1 & 57.2 \\
& AGDO Scheduler  & \underline{40.6} & 45.7 & 5.5 & 20.7  & 62.3 & 65.6 & 37.3 & 53.6  \\
& Suffix Anchor & \underline{40.6} & 42.2 & \underline{33.2} & \textbf{45.7} & 57.6 & 60.4 & \textbf{58.1} & \underline{59.1} \\
\midrule
\multirow{4}{*}{Proposed} & Attn (u) & 17.2 & 21.5 & 1.6 & 9.8  & 44.8 & 55.7 & 22.7 & 31.7 \\
& Attn (m) & 14.8 & 25.4 & 6.6 & 7.8 & 55.6 & 61.9 & 25.6 & 49.3 \\
& Attn (l) & 13.3 & 23.0 & 0.4 & 3.9 & 57.9 & 59.7 & 28.9 & 48.9 \\
& Evolution (393 params) & \textbf{50.0} & \textbf{53.9} & \textbf{35.2} & \underline{40.6} & \textbf{65.5} & \underline{65.9} & \underline{54.1} & \textbf{66.5} \\
\bottomrule
\end{tabular}
\label{tab:llada}
\end{table*}

\section{Experiments}
\subsection{Experiment Setup}

To validate the efficiency and generalization capacity of the proposed method, we evaluate our scheduler on two popular dLLMs: LLaDA-8B-Instruct \cite{nie2026large} and Dream-7B-Instruct \cite{ye2025dream}. We train the scheduler and evaluate it on four reasoning and planning tasks spanning mathematics and logic: GSM8K \cite{cobbe2021training}, Math \cite{hendrycks2measuring}, Countdown \cite{tinyzero}, and StrategyQA \cite{geva2021did}. For all datasets, we follow the standard training and test splits. For Math, we use the Math-500 subset as the test set for efficient evaluation. Detailed introductions to these datasets are provided in Appendix~\ref{sec:data}.

To demonstrate the superiority of our method, we compare against several classical and competitive fixed-budget baselines. In addition to conventional heuristic and Block-AR schedulers, we include the following recent approaches: \textbf{(i) Early Decision Matters (EDM, 5M parameters)} \citep{kim2026early}, which trains a scorer to guide the initial denoising steps. Since its dataset construction is nontrivial, we adopt the same evaluation protocol and report the results from the original paper~\citep{kim2026early} for fair comparison; \textbf{(ii) CCD} \citep{chen2025beyond}, which leverages historical average confidence scores for more consistent guidance; \textbf{(iii) AGDO Scheduler} \citep{deng2026beyond}, which first selects candidates based on last-layer valid attention scores and then determines which tokens to denoise using Top-1 probability; and \textbf{(iv) Suffix Anchor} \citep{park2026confidence}, which appends a fixed prompt at the end of the generation region to mitigate EOS overflow. We exclude two neural schedulers from comparison: \citet{jazbec2025learning} adopts adaptive denoising speed, making it incompatible with our fixed-budget setting; and UPO~\citep{hong2026improving} supports only one token per step, which is mismatched with our fast-denoising protocol.

To ensure robustness, we evaluate under multiple configurations with generation length $L \in \{128, 256\}$ and denoising budget $T \in \{16, 32\}$. As noted in prior work~\cite{ben2026accelerated,jazbec2025learning,luxembourg2025plan}, performance gaps between schedulers diminish with large budgets, and pure AR decoding can be competitive~\cite{ni2026flexibility}. We therefore focus on the more challenging fast-denoising regime, which better aligns with the parallel decoding strength of dLLMs. For CMA-ES optimization, we use a batch size of 100, a maximum of 10 generations, and an initial step size of 0.2, with all other parameters set to default values in the pycma package~\cite{hansen2019pycma}. 

\subsection{Experiment Results} \label{sec:experiment_result}



The results on LLaDA-8B-Instruct are presented in Table~\ref{tab:llada}, while those on Dream-7B-Instruct are deferred to Appendix~\ref{sec:dream}, as they lead to similar conclusions. Overall, our proposed evolutionary heuristic scheduler demonstrates consistently strong performance, outperforming all baselines in the majority of settings. The advantage of our method becomes particularly pronounced on more challenging tasks and under tighter budgets. For instance, on MATH with the most challenging configuration ($T=16$, $L=256$, i.e., 16 tokens denoised per step), our scheduler achieves 25.0\% accuracy, substantially outperforming the best baseline at 17.2\%. On Countdown, our method is the only one that consistently exceeds 35\% accuracy across settings, while most baselines remain below 10\% when $T=16$ and $L=256$.

Moreover, we observe that attention-based heuristics exhibit inconsistent performance. Middle-layer attention scores generally outperform those from upper or lower layers, aligning with recent findings that the middle layers of Transformers serve as a "global workspace" for reasoning \cite{gurnee2026verbalizable}. However, valid attention scores alone are far from sufficient, as they frequently underperform simple confidence-based heuristics. This observation reinforces our earlier claim (Fig.~\ref{fig:analysis}) that attention signals offer complementary rather than competitive value: they excel in certain niche scenarios but require combination with other heuristics to achieve robust performance.

\subsection{Ablation and Generalization Study}

\begin{table}[t!]
\centering
\caption{Ablation and Generalization Experiment of LLaDA-8B-Instruct \cite{nie2026large} in Math \cite{hendrycks2measuring}}
\centering
\begin{tabular}{lcccc}
\toprule
\multirow{2}{*}{Scheduler} & \multicolumn{2}{c}{L=128} & \multicolumn{2}{c}{L=256} \\
\cmidrule(lr){2-3} \cmidrule(lr){4-5} 
& T=16 & T=32 & T=16 & T=32  \\
\midrule
Proposed & \underline{21.4} & \textbf{27.2} & \textbf{25.0} & \textbf{27.8} \\
\midrule
w/o MF & 21.0 & \underline{25.8} & 19.6 & 24.0 \\
w/o skip & \underline{21.4} & 24.0 & 20.0 & 23.4 \\
\midrule
GSM8K & \textbf{22.6} & 24.8 & 20.6 & \underline{24.6} \\
Countdown & 14.0 & 18.0 & 20.4 & 15.8 \\
StrategyQA & 14.2 & 22.0 & 18.2 & 16.6 \\
\midrule
Dream-7B-Instruct & 20.4 & 23.4 & 22.0 & 21.2 \\
\midrule
L=128 & -- & -- & 21.4 & 24.4 \\
L=256 & 20.6 & 24.2 & -- & -- \\
T=16 & -- & 24.4 & -- & 24.6 \\
T=32 & 20.0 & -- & \underline{22.6} & -- \\
\bottomrule
\end{tabular}
\label{tab:expand}
\end{table}

In this section, we take LLaDA-8B-Instruct \cite{nie2026large} and the Math dataset \cite{hendrycks2measuring} as a case study to conduct ablation and generalization analyses. The core design of our method lies in the network architecture; accordingly, we evaluate the impact of removing the mean-field embedding and the skip linear layer. To assess generalization capacity, we test the scheduler trained on different source datasets (GSM8K, Countdown, and StrategyQA), different backbone models (Dream-7B-Instruct), and different inference settings (generation length $L$ and denoising budget $T$). Since exhaustively evaluating all combinations across these dimensions would be computationally prohibitive, we select the most challenging dataset, Math, to make the conclusions more compelling and representative. Additionally, a comparison between methods under higher decoding budgets is provided in Appendix~\ref{sec:dream}.

The experimental results are presented in Table~\ref{tab:expand}. For the ablation study, we observe that removing either the mean-field embedding (w/o MF) or the skip linear layer (w/o skip) leads to performance degradation to varying extents. Notably, the degradation patterns differ across configurations. Eliminating mean-field embedding suffers more severely when the sequence length is larger, aligning with our expectation that longer sequences require stronger global contextual aggregation, which the mean-field embedding provides. Furthermore, the skip linear layer provides effective support for zero-order optimization training by reducing the difficulty of optimizing a purely non-linear structure.


Regarding the generalization experiments, transferring the scheduler to different datasets leads to varied degradation. The GSM8K-trained scheduler performs best among cross-dataset transfers, even slightly outperforming the Math-trained one under $T=16, L=128$, which is expected given the distributional similarity between the two math reasoning datasets. In contrast, schedulers trained on Countdown or StrategyQA suffer more substantial drops, suggesting that the scheduler learns domain-specific priors rather than purely universal heuristics. Nevertheless, since the scheduler inputs consist of effective manually designed heuristics, its performance remains at least comparable to those heuristic-based schedulers, avoiding complete failure even under domain shifts. When transferring to a different backbone (Dream-7B-Instruct) or unseen inference configurations (length and budget), we observe milder degradation, with results remaining highly competitive against prior work (Table~\ref{tab:llada}). Notably, the degradation from cross-dataset shifts is consistently larger than that from cross-configuration shifts, indicating that the scheduler captures a combination of domain-specific scheduling priors and universal heuristics. This suggests that mixed-dataset training could be a promising direction to improve robustness, which we leave for future work.

\section{Conclusion}

In this paper, we analyze the failure modes of confidence-based heuristic schedulers in dLLMs. Through attention-based analysis, we reveal that EOS Overflow and Proximal Bias arise from excessive attention to invalid tokens such as \text{[MASK]} and \text{[EOS]}. Building on this insight, we show that valid attention scores serve as a strong complementary signal, yet no single heuristic suffices due to the highly context-dependent nature of optimal denoising trajectories. To address this challenge, we propose a lightweight evolutionary heuristic scheduler optimized via CMA-ES. By dynamically integrating multiple heuristics with a mean-field contextual embedding, our method achieves strong performance using only 393 trainable parameters, serving as the most parameter-efficient neural scheduler for dLLMs to date. Experiments across four challenging reasoning and planning tasks on LLaDA and Dream demonstrate consistent gains over strong baselines, especially in difficult settings with limited denoising budgets. While the learned scheduler exhibits some domain specificity, this opens promising avenues for future work on mixed-dataset training and adaptive denoising speed designs to further improve generalization and efficiency.





\bibliography{aaai2026}


\clearpage

\section{Appendix}
\appendix


\subsection{Literature Review} \label{sec:related_work}
Early work, inspired by the concept of confidence in AR LLMs \cite{kang2026scalable}, designs denoising schedulers under the assumption that tokens with higher confidence are more likely to be correctly predicted, and that revealing them first can provide reliable guidance for subsequent tokens. Metrics such as top-1 probability \cite{wu2025fast}, entropy \cite{ben2026accelerated}, and margin probability \cite{kim2025train} have been introduced and shown to outperform random denoising \cite{zheng2025masked}. Building on these, subsequent methods propose further empirical improvements. For instance, \citep{chen2025beyond} observed inconsistency in individual token predictions and proposed CCD, which uses historical average confidence as a more stable denoising metric. \citep{cao2026search} noted that average confidence over the entire denoising process correlates with final accuracy and introduced PBS, which applies beam search to retain trajectories with the highest cumulative confidence. To address the EOS Overflow issue, \citep{nie2026large} and \citep{park2026confidence} proposed directly suppressing [EOS] tokens or down-weighting confidence near sentence endings. However, these approaches may result in overly long outputs and lack flexibility.

To prevent [EOS] tokens from appearing too early and interfering with generation, Block-AR has emerged as an alternative paradigm \cite{arriola2025block}. By allowing arbitrary diffusion within blocks while decoding blocks auto-regressively, Block-AR achieves better contextual consistency, yet introduces new challenges. For example, \citep{lu2026adablockdLLM} and \citep{zhang2026swordsman} observed that block size is a critical hyper-parameter affecting generation quality, and more importantly, splitting a sentence or semantic paragraph across blocks can cause significant performance degradation. To address this, \citep{lu2026adablockdLLM} and \citep{zhang2026swordsman} proposed AdaBlock and Swordsman, respectively, which adaptively determine block boundaries based on confidence-related metrics, following empirical heuristics.  Moreover, \citep{kim2026early, ye2025beyond} also argue that reintroducing AR constraints in dLLMs could limit their potential for complex reasoning and planning tasks.

As noted by \citep{chen2025beyond}, most of the above methods rely on empirical heuristics with limited theoretical guarantees. As a result, another line of work aims to train a neural scheduler jointly with the dLLM backbone to learn optimal denoising trajectories by RL or from offline data. For instance, \citep{huang2026reinforcing} proposed LLaDOU, which first formulates denoising as a Markov Decision Process (MDP) and employs the Plackett–Luce model to model token selection probabilities at each denoising step. They further introduced RemeDi, incorporating a pre-training warmup with a remasking mechanism \cite{huang2026dont} to improve performance. To reduce training costs, recent efforts focus on training the scheduler while keeping the dLLM backbone frozen, which also achieves competitive results. For example, \citep{hong2026improving} and \citep{jazbec2025learning} both adopt GRPO \cite{shao2024deepseekmath} to train a Transformer-based scheduler, taking as input either the hidden features from the dLLM or the prediction probabilities of each token, respectively. Meanwhile, \citep{kim2026early} train a Transformer to predict whether an initial trajectory is likely to lead to a correct answer using offline data, and then use it to score and guide early inference steps. Overall, these methods require Transformers with parameter counts ranging from 300K to 134M, incurring additional training and inference costs. In contrast, our proposed scheduler is a carefully designed network with only 393 parameters, making it, to the best of our knowledge, the most lightweight neural scheduler to date.

\subsection{Training Algorithm} \label{sec:algorithm}
The detailed training process of our neural scorer is provided at Algorithm \ref{alg:cmaes}.

\begin{algorithm}[htbp]
\caption{CMA-ES Training for the Neural Scorer}
\label{alg:cmaes}
\begin{algorithmic}[1]
\Require Model \(\mathrm{h}_{\Theta}(\cdot)\) with parameters \(\Theta \in \mathbb{R}^D\); training set \(\mathcal{D}_{\text{train}}\); population size \(\lambda\); initial step size \(\sigma_0\); batch size \(B\); maximum generations \(G_{\max}\).
\Ensure Optimized parameters \(\Theta^*\).

\State Initialize \(\Theta_0\) randomly (e.g., Xavier uniform).
\State Initialize CMA-ES parameters: mean \(\mathbf{m} \leftarrow \Theta_0\), step size \(\sigma \leftarrow \sigma_0\), covariance \(\mathbf{C} \leftarrow \mathbf{I}\).
\State \(g \leftarrow 0\).

\While{\(g < G_{\max}\) and not converged}
    \State Sample \(\lambda\) candidate solutions \(\{\Theta_i\}_{i=1}^\lambda \sim \mathcal{N}(\mathbf{m}, \sigma^2 \mathbf{C})\).
    \State Sample a fixed mini-batch \(\mathcal{B} \subset \mathcal{D}_{\text{train}}\) of size \(B\).
    
    \For{\(i = 1\) to \(\lambda\)}
        \State Assign \(\Theta_i\) to scorer \(\mathrm{h}_{\Theta_i}\).
        \State Generate answers for all questions in \(\mathcal{B}\) using \(\mathrm{h}_{\Theta_i}\).
        \State Compute fitness \(f_i \gets \frac{1}{B} \sum_{q \in \mathcal{B}} \mathbf{1}\{\text{correct}(q)\}\).
    \EndFor
    
    \State Sort candidates by fitness: \(f_{(1)} \ge \cdots \ge f_{(\lambda)}\).
    \State Select top \(\mu\) candidates (\(\mu = \lambda/2\)).
    \State Update mean: \(\mathbf{m} \leftarrow \sum_{i=1}^\mu w_i \Theta_{(i)}\), with weights \(w_i = \frac{\log(\mu+1)-\log(i)}{\sum_{j=1}^\mu (\log(\mu+1)-\log(j))}\).
    \State Update covariance matrix \(\mathbf{C}\) and step size \(\sigma\) using standard CMA-ES rules.
    \State \(g \leftarrow g + 1\).
\EndWhile

\State \Return \(\Theta^* \gets \mathbf{m}\).
\end{algorithmic}
\end{algorithm}

\begin{table*}[t!]
\centering
\caption{Model Performance on Dream-7B-Instruct \cite{ye2025dream}. For Dream, which is based on an AR LLM backbone, the attention scores are shifted by one position to align with the token prediction order. Regarding EDM, the original paper does not report results for Countdown or StrategyQA on Dream.}
\begin{adjustbox}{width=0.8\textwidth}
\centering
\begin{tabular}{llcccccccc}
\toprule
& & \multicolumn{4}{c}{GSM8K} & \multicolumn{4}{c}{Math} \\
\cmidrule(lr){3-6} \cmidrule(lr){7-10}
\multirow{2}{*}{Type} & \multirow{2}{*}{Scheduler} & \multicolumn{2}{c}{L=128} & \multicolumn{2}{c}{L=256} & \multicolumn{2}{c}{L=128} & \multicolumn{2}{c}{L=256} \\
\cmidrule(lr){3-4} \cmidrule(lr){5-6} \cmidrule(lr){7-8} \cmidrule(lr){9-10}
& & T=16 & T=32 & T=16 & T=32 & T=16 & T=32 & T=16 & T=32 \\
\midrule
\multirow{3}{*}{Heuristics} & Top-1 Prob & 37.4 & 42.8 &22.0 & 45.8 & 15.8 & 19.4 & 10.8 &  18.2 \\
& Entropy & 26.2 & 38.9 & 16.1 & 39.7 & 13.4 & 19.0 & 5.8 & 17.8 \\
& Prob Margin & 38.3 & 45.4 & 26.8 & 44.6 & 14.6 & 16.8 & 11.8 & 17.6 \\
\midrule
\multirow{3}{*}{Block-AR (B=32)} & Top-1 Prob & 23.0 & 56.6 & 2.1 & 16.5 & 7.8 & 23.8 & 1.0 & 5.0  \\
& Entropy & 16.8 & 50.9 & 2.2 & 10.1 & 8.2 & 19.4 & 1.8 & 6.8   \\
& Prob Margin & 28.8 & \underline{58.8} & 2.0 & 20.8 & 11.6 & \underline{24.4} & 2.2 & 6.8 \\
\midrule
\multirow{4}{*}{Recent Works} & EDM (5M params) & -- & -- &  -- &   \underline{52.4} & -- & -- & -- &  \underline{21.4} \\
& CCD & 32.3 & 42.8 & 18.6 & 38.0 & 13.6 & 21.4 & 6.8 & 17.4 \\
& AGDO Scheduler & 36.5 & 41.6 & 20.2 & 39.9 & 16.0 & 19.8 & 11.0 & 17.6 \\
& Suffix Anchor & \underline{42.3} & 46.4 &  \underline{28.4} & 37.0 & \underline{19.4} & 22.2 & \textbf{17.0}  & 20.2 \\
\midrule
\multirow{4}{*}{Proposed} & Attn (u) & 16.7 & 48.1 & 1.6 & 12.6 & 4.6  & 19.4  & 2.0 & 6.8 \\
& Attn (m) & 37.4 & 42.8 &30.7 & 41.3 & 11.7 & 15.4 & 12.0 & 14.8 \\
& Attn (l) & 9.5 & 46.0 & 2.0 & 5.1 & 2.4 & 14.4 & 2.0 & 2.0 \\
& Evolution (393 params) & \textbf{43.5} & \textbf{58.9} & \textbf{37.5} & \textbf{60.3} & \textbf{27.2} & \textbf{27.0} & \underline{15.2} & \textbf{22.6} \\
\bottomrule
\toprule
& & \multicolumn{4}{c}{Countdown} & \multicolumn{4}{c}{StrategyQA} \\
\cmidrule(lr){3-6} \cmidrule(lr){7-10}
\multirow{2}{*}{Type} & \multirow{2}{*}{Scheduler} & \multicolumn{2}{c}{L=128} & \multicolumn{2}{c}{L=256} & \multicolumn{2}{c}{L=128} & \multicolumn{2}{c}{L=256} \\
\cmidrule(lr){3-4} \cmidrule(lr){5-6} \cmidrule(lr){7-8} \cmidrule(lr){9-10}
& & T=16 & T=32 & T=16 & T=32 & T=16 & T=32 & T=16 & T=32 \\
\midrule
\multirow{3}{*}{Heuristics} & Top-1 Prob & 31.3 & 36.7 & 16.0 & \underline{31.6} & 46.9 & 67.4 & 33.6 & 45.3 \\
& Entropy & 22.7 & 34.0 & \underline{18.4} & 29.7 & 44.0 & 60.3& 26.2& 42.6 \\
& Prob Margin & \underline{35.2} & 37.1 & 12.5 & 29.7 & 51.1 & 67.8 & 40.6 & 51.7 \\
\midrule
\multirow{3}{*}{Block-AR (B=32)} & Top-1 Prob & 30.5 & 29.3 & 0.0 & 17.2 & 49.1 & 68.5 & 14.4 & 43.5 \\
& Entropy &18.4  & 36.7 & 0.0 & 8.2 & 44.8 & 58.2  & 25.9 & 44.1 \\
& Prob Margin & 27.0 & 30.5 & 0.0 & 21.9 & 51.1 & \textbf{69.1} & 11.1 & 44.5 \\
\midrule
\multirow{3}{*}{Recent Works} & CCD & 25.4 & 34.4 & 12.1 & 25.6 & 42.1 & 62.0 & 29.0 & 42.4 \\
& AGDO Scheduler & 29.3 & \underline{40.2} & \underline{18.4} & 28.5 & 45.9 & 58.7 & 34.2 & 46.0 \\
& Suffix Anchor & 14.5 & 33.6 & 2.7 & 10.9 & \underline{61.7} & 64.5 & 53.7 & 61.3 \\
\midrule
\multirow{4}{*}{Proposed} & Attn (u) & 5.5 & 17.2 & 0.4 & 2.34 & 50.9 & 58.9 & 29.5 & 53.9 \\
& Attn (m) & 10.9 & 15.2 &  0.0 & 4.3 & 58.1 & 60.1 & \underline{61.4} & \underline{62.2} \\
& Attn (l) & 3.9 & 34.0 & 5.5 & 8.7 & 22.7 & 22.9 & 23.6 & 26.6 \\
& Evolution (393 params) & \textbf{47.3} & \textbf{48.4} & \textbf{19.9} & \textbf{33.6} & \textbf{63.8} & \underline{68.6} & \textbf{65.6} & \textbf{65.1} \\
\bottomrule
\end{tabular}
\end{adjustbox}
\label{tab:dream}
\end{table*}

\subsection{Dataset Introduction} \label{sec:data}
\begin{itemize}[left=0pt]
\item \textbf{GSM8K \cite{cobbe2021training}:} A widely used benchmark of grade-school math word problems that require multi-step arithmetic reasoning. Each problem is paired with a natural-language solution, making it a standard testbed for evaluating chain-of-thought reasoning.
\item \textbf{Math \cite{hendrycks2measuring}:} A challenging collection of competition-level mathematics problems covering topics such as algebra, geometry, and number theory. For efficient evaluation, we adopt the Math-500 subset as the test set.
\item \textbf{Countdown \cite{tinyzero}:} An arithmetic planning task in which the model must combine a given set of numbers with basic operations to reach a target value. Following \cite{kim2026early}, we adopt the same 1-shot prompting strategy to ensure valid outputs and fair comparison.
\item \textbf{StrategyQA \cite{geva2021did}:} A logical reasoning benchmark of open-domain yes/no questions that require implicit multi-hop reasoning, where the necessary reasoning steps are not explicitly stated and must be inferred by the model.
\end{itemize}

\subsection{Expanded Experiment Results} \label{sec:dream}

\begin{figure}[h!]
\centering
\includegraphics[width=0.45\textwidth]{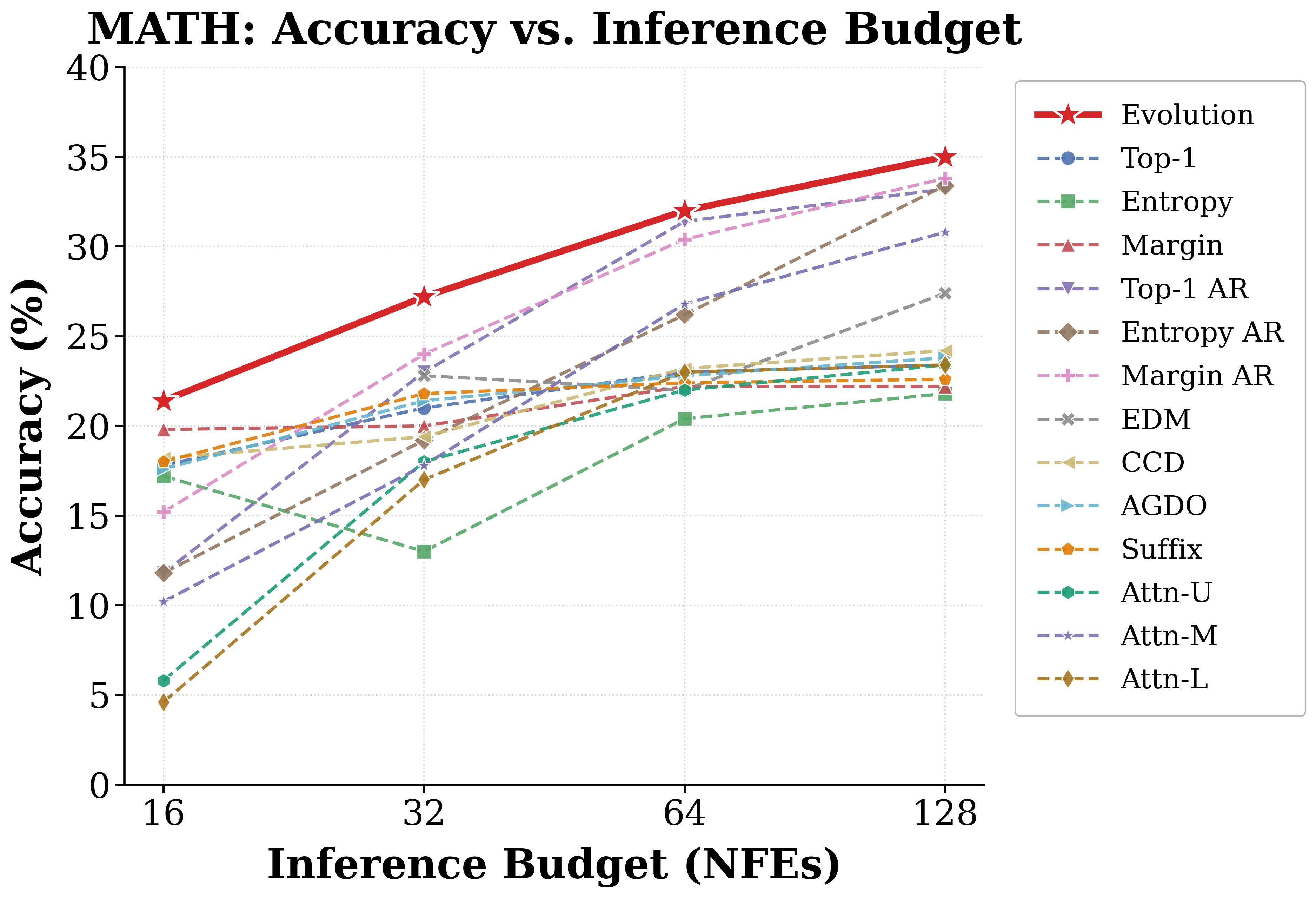}
\caption{Performance comparison across inference budgets: our method consistently outperforms baselines on Math500 using LLaDA ($L=128$).}
\label{fig:NFE}
\end{figure}

The experimental results for Dream are presented in Table~\ref{tab:dream} and lead to conclusions consistent with our analysis in Section~\ref{sec:experiment_result}. Furthermore, the results for LLaDA on Math under different inference budgets are shown in Fig.~\ref{fig:NFE}, where our proposed method consistently outperforms others across all settings.

\end{document}